\documentclass[sigconf,screen=true,anonymous=false,bookmarks=false]{acmart}
\iftrue
\usepackage{titlesec}
\titlespacing\section{2pt}{5pt plus 1pt minus 1pt}{0pt plus 1pt minus 1pt}
\titlespacing\subsection{2pt}{5pt plus 1pt minus 1pt}{0pt plus 1pt minus 1pt}
\titlespacing\subsubsection{2pt}{5pt plus 1pt minus 1pt}{2pt plus 1pt minus 1pt}
\usepackage[inline]{enumitem}
\setlist{leftmargin=5.08mm}
\fi

\iftrue
\fi

\usepackage{lipsum}
\usepackage{graphicx}
\usepackage{amsmath}
\usepackage{footnote}
\usepackage{mathtools}
\usepackage{mathrsfs}     
\usepackage{comment}
\usepackage[subrefformat=parens,labelformat=parens]{subfig}
\usepackage{bm,bbm}
\usepackage{multirow}
\usepackage{threeparttable,booktabs}
\usepackage{blkarray}
\usepackage{tikz}
\usetikzlibrary{positioning,calc,fit,decorations.pathmorphing,shapes.geometric,shapes.gates.logic.US,calc}
\usepackage{tikz-3dplot}
\usepackage{balance}
\usepackage{courier}                                       
\usepackage{cleveref}                                      
\usepackage[mathcal]{eucal}
\usepackage[]{algpseudocode}                               
\algrenewcommand\textproc{\texttt}
\makeatletter\let\float@addtolists\relax\makeatother
\usepackage{algorithm}
\usepackage{filecontents}                                  
\usepackage{pgfplots}
\usepackage{pgfplotstable}
\usepackage{pgf-pie}
\usepgfplotslibrary{groupplots}
\pgfplotsset{compat=newest}
\usepackage[figuresright]{rotating}
\usepackage{xcolor,colortbl}                               

\renewcommand{\vec}[1]{\boldsymbol{#1}}

\newcommand{\minisection}[1]{\vspace{.06in}\noindent{\textbf{#1}}}

\theoremstyle{plain}

\theoremstyle{definition}

\algrenewcommand\textproc{\texttt}

\usepackage[skip=1pt]{caption}            
\definecolor{CUHKorange}{RGB}{244,106,18} 
\definecolor{CUHKblue}{RGB}{0,111,190}    
\definecolor{CUHKgreen}{RGB}{0,127,128}   
\definecolor{CUHKred}{RGB}{228,46,36}     
\definecolor{CUHKyellow}{RGB}{198,148,34} 
\definecolor{CUHKdark}{RGB}{114,44,114}   
\definecolor{CUHKmiddle}{RGB}{144,44,144} 

\usepackage{tcolorbox}
\tcbuselibrary{skins,breakable}
    {\endtcolorbox}

\graphicspath{{./figs/}}
\setcopyright{acmcopyright}
\let\vec\mathbf
\usepackage{multirow}

\definecolor{myorange}{RGB}{238,97,42}  %
\definecolor{myblue}{RGB}{40,120,181}  
\definecolor{mygrey}{RGB}{166,166,166}  %
\definecolor{mygreen}{RGB}{180,210,36}  
\definecolor{myred}{RGB}{238,0,0}       
\definecolor{myyellow}{RGB}{198,148,34} 
\definecolor{mydark}{RGB}{114,44,114}   
\definecolor{mymiddle}{RGB}{144,44,144} 
\definecolor{mylight}{RGB}{167,44,167}  
\definecolor{myblue1}{RGB}{137,157,192}  
\definecolor{mygreen1}{RGB}{69,137,148}  

\definecolor{blockblue}{HTML}{6E9ECE}
\definecolor{blockyellow}{HTML}{EFDBB9}
\definecolor{blockgreen}{HTML}{269464}
\definecolor{blockorange}{HTML}{CC5618}
\definecolor{blockgray}{HTML}{A19897}
\definecolor{blockred}{HTML}{E6928F}
\definecolor{blockpurple}{HTML}{A264A2}
\definecolor{blockbrown}{HTML}{84574d}

\begin{document}
\date{}

\title{MixPE: Quantization and Hardware Co-design for Efficient LLM Inference}

\author{Yu Zhang}       \affiliation{\institution{The Chinese University of Hong Kong}}  
\author{Mingzi Wang}  \affiliation{\institution{Tsinghua University}}
\author{Lancheng Zou}      \affiliation{\institution{The Chinese University of Hong Kong}}
\author{Wulong Liu}    \affiliation{\institution{Huawei Noah's Ark Lab}}
\author{Hui-Ling Zhen}      \affiliation{\institution{Huawei Noah's Ark Lab}}
\author{Mingxuan Yuan} \affiliation{\institution{Huawei Noah's Ark Lab}}
\author{Bei Yu}        \affiliation{\institution{The Chinese University of Hong Kong}}
\begin{abstract}
    Transformer-based large language models (LLMs) have achieved remarkable success as model sizes continue to grow, yet their deployment remains challenging due to significant computational and memory demands. Quantization has emerged as a promising solution, and state-of-the-art quantization algorithms for LLMs introduce the need for mixed-precision matrix multiplication (mpGEMM), where lower-precision weights are multiplied with higher-precision activations. Despite its benefits, current hardware accelerators such as GPUs and TPUs lack native support for efficient mpGEMM, leading to inefficient dequantization operations in the main sequential loop.

    To address this limitation, we introduce MixPE, a specialized mixed-precision processing element designed for efficient low-bit quantization in LLM inference. MixPE leverages two key innovations to minimize dequantization overhead and unlock the full potential of low-bit quantization. First, recognizing that scale and zero point are shared within each quantization group, we propose performing dequantization after per-group mpGEMM, significantly reducing dequantization overhead. Second, instead of relying on conventional multipliers, MixPE utilizes efficient shift\&add operations for multiplication, optimizing both computation and energy efficiency. Our experimental results demonstrate that MixPE surpasses the state-of-the-art quantization accelerators by $2.6\times$ speedup and $1.4\times$ energy reduction.
\end{abstract}

\maketitle
\pagestyle{empty}

\section{Introduction}
\label{sec:intro}

Large language models have sparked a new revolution across a broad spectrum of tasks, exerting a profound influence on our daily lives.
However, the colossal size of LLMs results in high computation and energy costs to train and serve these models.

\begin{figure}
    \centering
    \includegraphics[width=0.49\linewidth]{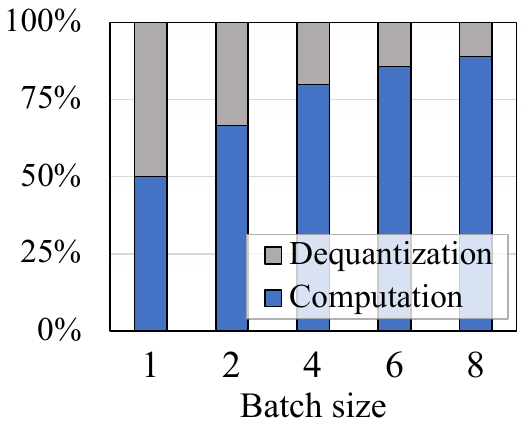}
    \includegraphics[width=0.49\linewidth]{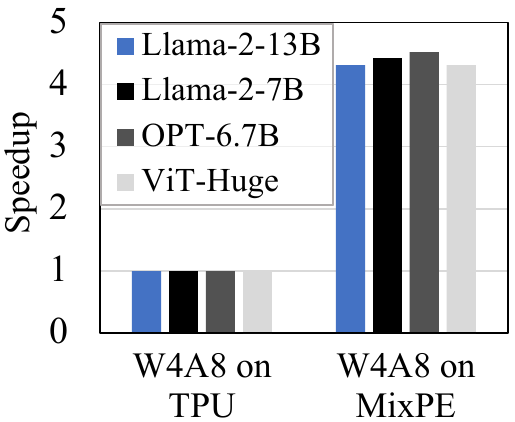}
    \caption{(Left) The dequantization overhead of Llama-2-7B quantized in W4A8. (Right) MixPE achieves over $4\times$ speedup when running LLMs compared to INT8-based TPUs.}
    \label{fig:intro}
\end{figure}

The AI industry is applying many techniques to reduce the cost of models. Quantization is one critical instance of these techniques, in which individual tensor values are cast from full-precision FP32 to a cheaper numeric standard. The most popular quantization formats include INT4, INT8, and FP16, given their vast hardware support. Based on these data formats, state-of-the-art integer quantization algorithms can be divided into three categories: 8-bit weight and 8-bit activation (W8A8), 4-bit weight and 16-bit activation (W4A16), 4-bit weight and 8-bit activation (W4A8). The later two mixed-precision methods, i.e., quantizing LLMs with low-precision weights and high-precision activations, have become particularly attractive as they save memory and computation costs while maintaining model accuracy~\cite{lin2024qserve, TRT}. This is because activations are harder to quantize than weights. More specifically, the weight distribution is quite uniform and flat, which is relatively easy to quantize. Previous work has shown that quantizing the weights of LLMs with 8-bit, 4-bit or even with 1-bit does not significantly degrade accuracy~\cite{dettmers2022int8, yao2022zeroquant, ma20241bit}. Conversely, activations are generated on-the-fly with a high variance, noticeably presented as dynamic outliers~\cite{xiao2023smoothquant, guo2023olive, lin2024awq}. These outliers can lead to significant accuracy degradation. 

Mixed-precision quantization shifts the key computation pattern of LLM inference from conventional General Matrix Multiplication (GEMM) to mixed-precision GEMM (mpGEMM), where the weight matrix is in lower precision (e.g., INT4) and the activation matrix remains in higher precision (e.g., INT8/FP16).
However, current hardware, such as GPUs and TPUs, does not natively support mpGEMM. Consequently, low-bit LLM inference systems have to implement inefficient dequantization-based approaches for mpGEMM~\cite{lin2024qserve, TRT}. Specifically, low-bit weights are dequantized to high precision before performing GEMM. Such extra operations are performed in the main loop and can become a performance bottleneck. When batch size is relatively small ($\leq 4$), dequantization overhead can account for over 20\% of the overall computation, as shown in the left figure of \Cref{fig:intro}. Meanwhile, as the GEMM is still computed in high precision, dequantization-based mpGEMM cannot take full advantage of low-precision computation.

Therefore, reducing dequantization overhead is crucial for achieving optimal throughput in LLM inference. Motivated by this, we introduce a \underline{Mix}ed-precision \underline{P}rocessing \underline{E}lement, MixPE, to accelerate low-bit LLM inference with a dedicated mpGEMM solution. 
Unlike conventional dequantization-based approaches, MixPE performs direct mixed-precision GEMM and postpones dequantization to after the per-group GEMM computation. This approach leverages the fact that the scale factor and zero point are shared within each quantization group, allowing for more efficient handling of low-precision weights. Additionally, to fully harness the benefits of low-precision computation, MixPE replaces traditional multipliers with efficient, low-power shift\&add operations for multiplying low-bit weights with high-bit activations.
Through holistically co-designing software (dequantization after mpGEMM) and hardware, i.e., shift\&add-based processing element (PE), MixPE not only reduces dequantization overhead but also exploits low-precision arithmetic for achieving high-throughput, energy-efficient LLM inference. Compared to conventional INT8-based TPUs, MixPE can achieve over $4\times$ speedup (as shown in the right figure in \Cref{fig:intro}) on LLM inference. Moreover, we present a parameterized design space exploration (DSE) framework that enables the evaluation of various GEMM accelerators. This framework allows us to map out the Pareto frontier, highlighting the optimal trade-offs between numerical fidelity and hardware efficiency. Our key contributions are as follows:
\begin{itemize}
    \item We introduce MixPE, a novel hardware accelerator that efficiently handles mixed-precision GEMM with minimal dequantization overhead by co-optimizing the quantization scheme and PE design. 
    \item We present a DSE framework that formalizes the design space as a trade-off between numerical fidelity and hardware cost. Through an exhaustive sweep of design variables, we identify MixPE as the Pareto frontier of numerical accuracy and performance.
    \item Experimental results demonstrate that, with W4A8 quantization, MixPE achieves a $2.6\times$ speedup and $1.4\times$ energy savings compared to state-of-the-art quantization accelerators. Additionally, for W4A16, MixPE delivers a $2.44\times$ speedup while reducing energy consumption by 68\% compared to traditional FP16 multiplication PEs.
\end{itemize}

\section{Related Work}
\label{sec:related}

\subsection{Integer Quantization in LLMs}

Nowadays, there is a rapid growth in the scale of large language models, which in turn requires significant hardware resources. For example, Llama-3-70B~\cite{dubey2024llama3} consumes approximately 148GB of memory just to hold its model weights (in FP16), far exceeding the capacity of a modern GPU like NVIDIA A100 or H100. This imposes a considerable challenge for LLM deployment. To reduce inference costs in LLM deployment, low-bit integer quantization, which maps floating point tensors to discrete level, has become a popular approach~\cite{guo2023olive, zhao2024atom, lin2024qserve}. Given n bits to represent the integer, the quantization process can be formulated as:
\begin{equation}
    Q_x = \lceil \frac{x}{s} +z \rfloor, \; s = \frac{x_{\max}-x_{\min}}{2^n-1}, \; z=\lceil -2^{n-1}-\frac{x_{\min}}{s} \rfloor,
\end{equation}
where $x$ is the floating-point value, $q_x$ is the n-bit quantized counterpart, $s$ is the scaling factor and $z$ is the zero point. Therefore, the dequantization process can be represented as:

\begin{equation}
    \hat{x} = (Q_x - z)\cdot s. \label{eq:dequant}
\end{equation}

For integer quantization, INT4 and INT8 formats are prevalently utilized in low-bit LLM inference. In this paper, we denote x-bit weight and y-bit activation quantization in LLMs as WxAy for abbreviation. Apart from bit precision, quantization can be achieved with different granularity, resulting in different trade-offs between accuracy and efficiency. For \textit{per-tensor} quantization, the entire tensor shares one set of scale and zero-point~\cite{nagel2021white}. Denoting channel as the last dimension of the input matrix, \textit{per-channel} quantization for weights or \textit{per-token} quantization for activations means that $s$ and $z$ are shared within each row of tensor~\cite{xiao2023smoothquant}. Dividing each channel into several sub-groups (g columns), \textit{per-group} quantization further reduces the degree of parameter sharing by using different $s$ and $z$ for every group within each row~\cite{lin2024awq}. The finer the granularity, the more precise the quantization, but the higher the overhead. Additionally, group quantization is often employed to improve the accuracy of 4-bit quantization. A typical group size is 128~\cite{zhao2024atom,lin2024awq,lin2024qserve}, where each contiguous block of 128 elements is quantized as a single group, allowing for more accurate representation of the data within each group.

\begin{figure}
    \centering
    \includegraphics[width=0.85\linewidth]{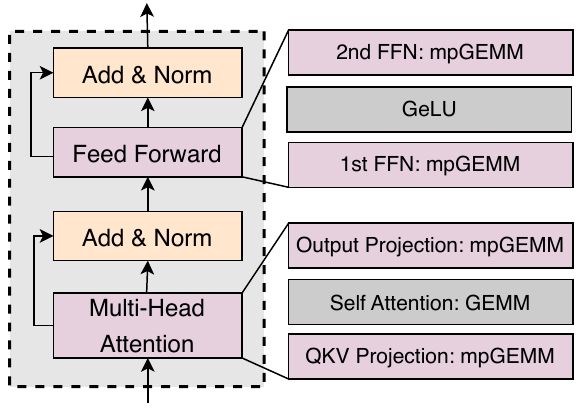}
    \caption{The mpGEMM operations in decoder-only LLMs.}
    \label{fig:mp}
\end{figure}

\subsection{mpGEMM in Low-bit LLM inference}

In decoder-only LLMs, using different bit-widths for weights and activations creates a unique requirement for mixed-precision General Matrix Multiplication, where the weight matrix uses lower precision while the activation matrix remains in higher precision. This mixed-precision requirement applies to primary GEMM operations in both multi-head attention and feed-forward blocks when weight and activation quantization bits differ (as illustrated in \Cref{fig:mp}).
However, current commercial hardware like GPUs and TPUs primarily support canonical GEMM, where both inputs are in the same format and bit-width, and lack native support for mpGEMM. Existing mpGEMM methods generally fall into two main categories: indirect mpGEMM, primarily dequantization-based mpGEMM, and direct mpGEMM, which includes lookup table (LUT)-based mpGEMM and processing element (PE)-based mpGEMM approaches.

\minisection{Indirect mpGEMM}
Dequantization-based mpGEMM first upscale the low-precision weights to match the high-precision activations and then perform conventional GEMM~\cite{TRT, lin2024awq, lin2024qserve}. Fig illustrates a dequantization-based mpGEMM example, where INT4 weights are multiplied by FP16 activations. While this approach supports various precision combinations, dequantization introduces additional operations that may become a performance bottleneck. Moreover, since the GEMM is executed in high precision, dequantization-based mpGEMM cannot fully exploit the benefits of low-precision computation.

\minisection{Direct mpGEMM}
LUT-based mpGEMM is one direct mpGEMM approach that uses lookup tables (LUTs) to implement mpGEMM~\cite{park2022lut, maleki2023look, mo2024lut}. Specifically, it precomputes dot products of high-precision activations with a limited set of low-precision weights and replaces the computation with simple lookups in the resulting table.
However, due to limited LUT support on existing LLM inference hardwares like GPUs, LUT-based mpGEMM approaches are often less effective than dequantizaiton-based approaches. Moreover, storing a lookup table for each potential weight-activation combination can require significant memory, especially as the range of data format increases. Another direct mpGEMM approach involves designing specialized processing elements (PEs) that can directly execute mixed-precision GEMM computations. BitFusion~\cite{sharma2018bit} proposes to leverage tensor bit-width adaptivity to reuse low-precision components (e.g., 4-bit integers) for higher-precision calculations (e.g., 8-bit integers) without additional overhead.
However, BitFusion is restricted to integer types, which limits its quantization efficiency and often results in increased memory and computation requirements. OLAccel~\cite{park2018ola} performs sparse high-precision computation for outliers in parallel with dense low-precision computation for regular tensors.
However, it stores tensors in off-chip memory, which results in longer memory access latencies and thus lower throughput. Both ANT~\cite{guo2022ant} and Olive~\cite{guo2023olive} propose novel data formats for efficiently storing large values within tensors.
However, these approaches require specialized data decoders, introducing additional area and computation overhead.
In contrast, our work, MixPE, introduces a flexible mpGEMM computation design that supports both integer and floating-point formats without introducing hardware overhead.

\section{Method}

\subsection{Motivation}

\begin{figure*}
    \centering
    \includegraphics[width=0.5\linewidth]{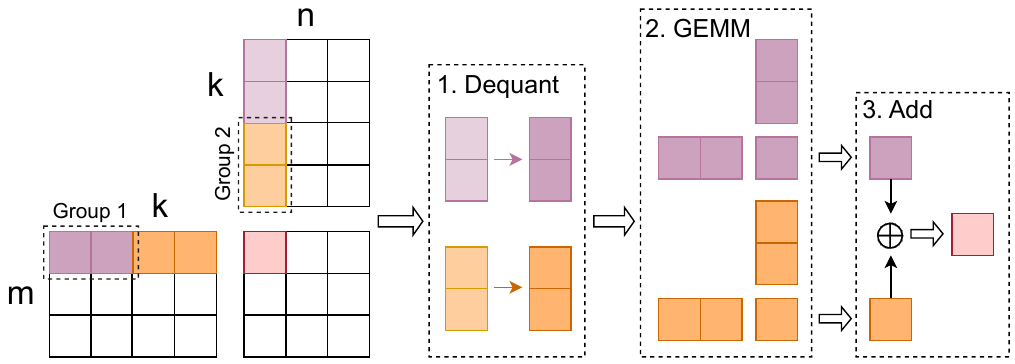}
     \includegraphics[width=0.49\linewidth]{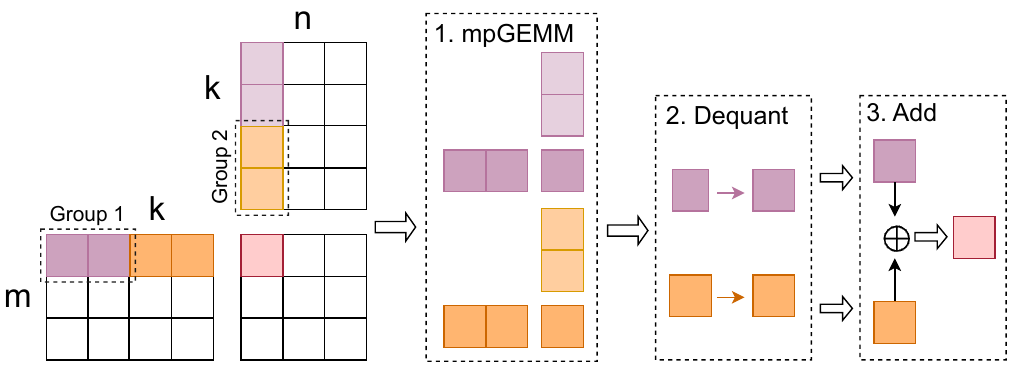}
    \caption{ \textbf{(Left)} Quantized GEMM on GPUs. The low-precision weights are first dequantized to high precision (Step \textcircled{1}). Each group then performs multiplication using conventional high-precision units (Step \textcircled{2}), and the results are accumulated to produce the final output (Step \textcircled{3}). \textbf{(Right)} Quantized mpGEMM on MixPE. The mixed-precision multiplication of each group is first computed by MixPE with efficient low-bit support(Step \textcircled{1}).
    The results are then dequantized and subsequently accumulated to produce the final output (Steps \textcircled{2},\textcircled{3}).}
    \label{fig:gemm}
\end{figure*}

Quantizing weights and activations can significantly reduce memory usage and improve computational throughput. Although model weights are known in advance and typically exhibit uniform distributions, achieving accurate representation remains challenging due to the limited precision of 4-bit integer formats. To further improve precision, group quantization is widely used. This method divides the weight matrix into subgroups and performs quantization within each group. To incorporate group quantization into the conventional GEMM pipeline, state-of-the-art quantization algorithms, including TensorRT-LLM-W4A16~\cite{TRT} and QServe-W4A8~\cite{lin2024qserve}, require dequantization operations in the main loop, as shown in the left figure of \Cref{fig:gemm}. For an $m\times n \times k$ quantized GEMM problem, $m$ represents the number of sequences (or batch size), while $n$ and $k$ correspond to channel dimensions. Both $m$ and $n$ are parallelizable dimensions, whereas $k$ serves as the reduction dimension, requiring a sequential main loop. The main loop includes over 100 iterations, which significantly dominate the runtime of the GEMM operation. 

The dequantization process in the main loop introduces two significant efficiency bottlenecks. First, dequantization requires additional multiplication and subtraction operations (see \Cref{eq:dequant}) on the large $n\times k$ weight matrix. In LLM inference, the batch size $m$ is typically small (e.g., 8, 16, 32), whereas $n$ and $k$ are much larger (e.g., 2048 or 4096). Upscaling such a large matrix during each iteration adds substantial overhead, quickly becoming a performance bottleneck. Second, the GEMM operations in the left figure of \Cref{fig:gemm} are still performed in high precision. This high-precision computation not only reduces the potential performance gains from low-bit quantization but also leads to increased power consumption and memory bandwidth usage.

We preview our group-quantized mpGEMM design in the right figure of \Cref{fig:gemm}. Observing each group shares a common set of scale and zero-point values, we propose performing mixed-precision GEMM first (Step \textcircled{1}), followed by per-group dequantization (Step \textcircled{2}). To enable this, we design a mixed-precision processing element optimized for efficient low-bit computation, unlocking the full potential of low-precision arithmetic. The dot product result for each group is then dequantized and accumulated to produce the final output (Step \textcircled{3}), thereby significantly reducing dequantization overhead in the main loop.

\subsection{Dequantize After GEMM}

Consider $k$-dimensional weight and activation vectors:
\begin{equation}
    \vec{w} = [w_0, w_1, \dots, w_{k-1}], \; \vec{x} = [x_0, x_1, \dots, x_{k-1}],
\end{equation}
where $\vec{w}$ is a row vector of the weight matrix $\vec{W_{n\times k}}$ and $\vec{x}$ is a row vector of the activation matrix $\vec{X_{m \times k}}$. Based on \Cref{eq:dequant}, their inner product can be expressed as:
\begin{equation}
    \vec{x} \vec{w}^\top \approx \sum_{i=0}^{k-1} x_i \cdot s_{w_i} (Q_{w_i} - z_{w_i}), \label{eq:gemm}
\end{equation}
where $s_{w_i}$ and $z_{w_i}$ are the scale and zero point of $w_i$, respectively. Here $x_i$ and $Q_{w_i}$ are in different precisions, with $Q_{w_i}$ in INT4 and $x_i$ in either INT8 or FP16. To perform GEMM on GPUs, $Q_{w_i}$ must be dequantized before being multiplied with $x_i$, corresponding to the left figure in \Cref{fig:gemm}.
In particular, we observe that, when group quantization is employed, the scale factor and zero-point are shared within each group. Let the number of groups be $N_{group} = k/g$, where $g$ is the group size, and denote the set of element indices in the $n$-th group as $G_n$. As a result, we can express the dot product in \Cref{eq:gemm} as follows:
\begin{align}
    \vec{x} \vec{w}^\top & \approx \sum_{n=0}^{N_{group}-1} s_{G_n} (\sum_{j \in G_n}^{} Q_{w_j} x_j - z_{G_n}\sum_{j \in G_n}^{} x_j), \label{eq:mpgemm} \\
    & \approx \sum_{n=0}^{N_{group}-1} (s_{G_n} \sum_{j \in G_n}^{} Q_{w_j} x_j - s_{G_n} z_{G_n}\sum_{j \in G_n}^{} x_j). \notag
\end{align}
Here $s_{G_n}$ and $z_{G_n}$ represent the scale factor and zero point of $G_n$, respectively.
From \Cref{eq:mpgemm}, we observe two key points. First, if multiplication between $Q_{w_j}$ and $x_j$ can be performed directly, dequantization can be applied after the inner-group dot product without sacrificing accuracy. This approach reduces the frequency of dequantization operations to $1/g$ of the original rate, significantly decreasing dequantization overhead in the main loop. Second, the computational complexity of the first term in \Cref{eq:mpgemm} is $O(k)$, while the second term is $O(1)$. As a result, most existing accelerators for Transformers~\cite{wang2021spatten, zou2024bie} prioritize accelerating the inner product term, with the remaining summation term delegated to software or other specialized computation units. Thus, designing an efficient computation engine for the mixed-precision dot product is essential to achieving optimal mpGEMM performance in LLM inference.

\subsection{Mixed-Precision Processing Elements} \label{sec:mixpe}

State-of-the-art quantization algorithms for LLM inference often quantize model weights to low-bit integer formats, achieving high theoretical throughput and a reduced memory footprint. In computer systems, integers are represented in binary with a fixed bit width, where each bit can be either 0 or 1. For a 4-bit integer (INT4), this allows for $2^4=16$ possible values, enabling INT4 to represent up to 16 distinct numbers. For W4A8 quantization, we employ an unsigned 4-bit (UINT4) quantization on weights as QServe~\cite{lin2024qserve}, and introduce a specialized mixed-precision processing element design tailored for quantized GEMM:
    \begin{align}
        & w (UINT4) \ast x (INT8) \notag \\
      & = (w_0 \ast 2^0 + w_1 \ast 2^1 + w_2 \ast 2^2 + w_3 \ast 2^3) \ast x  \notag \\
      & =  w_0 \ast x << 0 + w_1 \ast x<<1 + w_2 \ast x<<2 \notag \\
      & \quad + w_3 \ast x<<3 \notag  \\
      & = \sum_{i=0}^3 \mathbf{1}(w_i) (x<<i), \label{eq:pe} \\
       & \text{where} \; \mathbf{1}(w_i) 
        := \left\{
            \begin{aligned}
                1 , & \quad \text{the i-th bit of w is 1}, \\
                0 , & \quad \text{otherwise}, \notag
            \end{aligned}
            \right.
    \end{align}
where $w$ and $x$ represent the values in UINT4 weights and INT8 activations, respectively, and $<<$ denotes the left shift operator for integer data formats. Due to the structure of integer representation, original multiplications can be efficiently replaced by bit shifts. Specifically, shifting left by n bits is equivalent to multiplying by $2^n$. Based on \Cref{eq:pe}, we implement MixPE using shift and add operations, enabling highly efficient mixed-precision GEMM with weights in INT4 and activations in INT8. Theoretically, MixPE for W4A8 can achieve $2\times$ speed-up, 25\% - 50\% memory cost savings, and 25\% - 50\% communication savings, compared with the current INT8 multiplication module in tensor core, which is very promising for scaling-up next-generation hardware accelerators. Then, for W4A16 operations, even though bitwise shifts cannot replace multipliers as they are not applicable to floating-point values, efficient techniques still exist to achieve fast scaling by powers of two without the need for a traditional power-intensive multiplier. The MixPE design for W4A16 is detailed below:
    \begin{align}
        & w (UINT4) \ast x (FP16) \notag \\
      & = (w_0 \ast 2^0 + w_1 \ast 2^1 + w_2 \ast 2^2 + w_3 \ast 2^3) \ast x  \notag \\
      & = \sum_{i=0}^3 \mathbf{1}(w_i) (x \otimes 2^i), \label{eq:pefp} \\
       & \text{where} \; \mathbf{1}(w_i) 
        := \left\{
            \begin{aligned}
                1, & \quad \text{the i-th bit of w is 1}, \\
                0, & \quad \text{otherwise}. \notag
            \end{aligned}
            \right.
    \end{align}
where $\otimes$ represents a dedicated RTL module designed for efficient multiplication of floating-point values and powers of two. FP16 follows the IEEE 754~\cite{kahan1996ieee} standard, with a 1-5-10 bit format: 1 sign bit, 5 exponent bits, and 10 mantissa (fraction) bits, yielding an actual value of: value(FP16) = $(-1)^{\text{sign}} \times 2^{\text{E-bias}} \times (1.\text{mantissa})$. With this format, scaling an FP16 value by a power of two involves adjusting only the exponent bits, while the sign and mantissa remain unchanged. Specifically, we first perform a shift operation on the exponent bits, followed by bit-wise addition, enabling efficient power-of-two scaling. This process is similar to the INT4$\times$INT8 procedure, leveraging the same shift\&add approach for efficient computation. By directly modifying the exponent in FP16, we can scale activations by powers of two without traditional power-intensive multipliers. This design choice in MixPE for W4A16 significantly reduces computational overhead, minimizing operations and avoiding the latency and energy costs of conventional multipliers. 

\begin{figure}
    \centering
    \includegraphics[width=.98\linewidth]{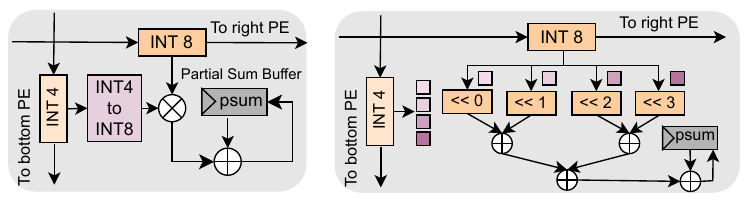}
    \caption{(Left) Traditional multiplier-based PE design.(Right) INT4$\times$INT8 processing element in our MixPE. MixPE achieves efficient computation by directly utilizing hardware-friendly shift operations and an optimized adder tree, eliminating the need to transform INT4 to INT8, leading to reduced power consumption and improved throughput. }
    \label{fig:pe}
\end{figure}

In the \Cref{fig:pe}, we provide a detailed comparison of the INT4$\times$INT8 weight-activation processing element between traditional multiplier-based designs and MixPE. The same topology applies to W4A16, where we also leverage shift\&add operations on the exponent bits of FP16 activations to achieve power-of-two scaling. In traditional multiplier-based PE, the INT4 activation must first be converted to INT8 before performing multiply-accumulate operations using standard multipliers (see the left figure in \Cref{fig:pe}), which fails to capitalize on the computational advantages of INT4. In contrast, MixPE achieves efficient computation between INT4 and INT8 by directly leveraging hardware-friendly shift operations and an optimized adder tree (see the right figure in \Cref{fig:pe}), significantly improving both power efficiency and throughput.

\begin{figure}[htbp]
    \centering
    \includegraphics[width=0.78\linewidth]{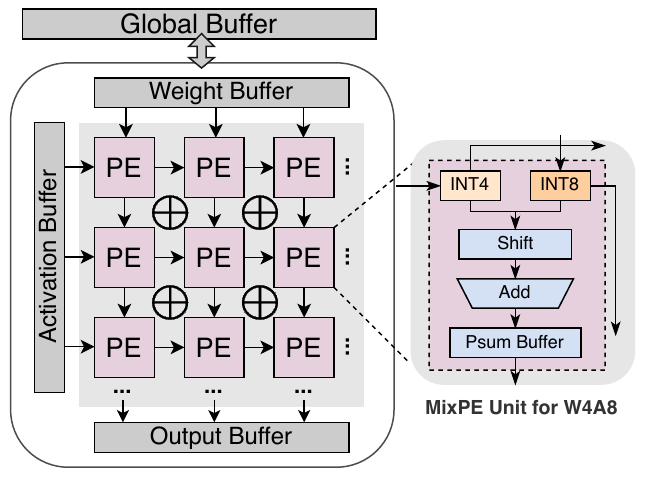}
    \caption{Architecture overview for integrating MixPE into the output stationary systolic array.}
    \label{fig:systolic}
\end{figure}

We then describe the architectural design for integrating MixPE into a systolic array, a structure also used in commercial deep learning accelerators like Google’s TPU~\cite{jouppi2017tpu}. Our design aligns with practical settings, where the weight tensor is quantized to low bit-width while the activation and output tensors retain higher precision. As such, we find that our design achieves the best benefits on the systolic array with the output-stationary dataflow~\cite{sharma2018bit}. The integration of MixPE into the output stationary systolic array architecture is illustrated in \Cref{fig:systolic}. This architecture, which communicates with memory via a global buffer, includes an input/weight buffer, multiple MixPE units, and an output buffer. The key innovation here is the MixPE design, which replaces traditional multipliers with efficient shifting and scaling modules. By incorporating MixPE, the systolic array achieves a hardware-efficient and energy-saving solution for mixed-precision GEMM computations, advancing quantization techniques and accelerating LLM inference tasks.

\subsection{Multi-Objective Design Space Exploration}

The primary goal of LLM quantization is to reduce memory usage and computational costs during inference. Quantization methods generally seek to balance two often competing factors: (1) quantization accuracy, which reflects the quality loss incurred by quantization relative to FP32 full precision, and (2) hardware efficiency, which combines computational efficiency (energy and silicon area for matrix multiplication) and memory efficiency (the number of bits needed to store or transfer a tensor). Popular quantization formats include INT4, INT8, and FP16, which form the basis of state-of-the-art quantization schemes such as W4A8, W8A8, and W4A16. This variety of data formats and configurations creates a complex design space, requiring careful trade-offs between accuracy, computational efficiency, and memory efficiency. Therefore, we first formalize the design space as a function of quantization accuracy and hardware cost. We then conduct an exhaustive exploration of design variables to identify Pareto-optimal design configurations.

\subsubsection{Quantization Accuracy}

First, to assess quantization accuracy, we calculate the quantization signal-to-noise ratio (SNR), a metric shown to reliably predict the end-to-end language model loss induced by quantization~\cite{darvish2023mx}. This approach enables extensive design space exploration without the need for costly experiments on each LLM quantization configuration. Specifically, SNR is defined as the ratio of the power of the original (non-quantized) signal to the power of the quantization noise:

\begin{equation}
    \text{SNR} := \frac{P_{\text{signal}}}{P_{\text{noise}}} = \log (\frac{\mathbf{E}[\|\vec{x} \|^2]}{\mathbf{E}[\|\vec{Q_x} - \vec{x}\|^2]}), \label{eq:snr}
\end{equation}
where $\vec{x} = [x_1, x_2, \dots, x_k] \in \mathbb{R}^k$ and $\| \vec{x}\|^2 = \sum_{i=1}^k x_i^2$. In quantization, the objective is to ensure that the quantized vector $\vec{Q_x}$ closely approximates the original vector $\vec{x}$, minimizing any discrepancies introduced by quantization. This goal corresponds to reducing the denominator term in \Cref{eq:snr}, which represents the quantization noise, thereby preserving the numerical fidelity of the original data. Generally, a higher SNR reflects a quantized vector that more accurately retains the direction and magnitude of the original non-quantized vector, denoting a high-quality quantization process. 

\subsubsection{Hardware Cost}

As shown in \Cref{fig:systolic}, we focus on systolic array architectures, a class of accelerators widely adopted in low-power TPUs~\cite{jouppi2017tpu}. Systolic array-based accelerators are highly configurable, encompassing a broad design space with trade-offs across performance, power consumption, circuit area, and memory cost. The accelerator design illustrated in \Cref{fig:systolic} features a 2D grid of parallel processing elements, arranged in a tiled structure. The count of PEs in each dimension (columns and rows) determines the aspect ratio of the chip, while the type of PEs is determined by the input data formats. 
Specifically, for conventional GEMM with the same input data format, we employ standard INT8$\times$INT8 PE and FP16$\times$FP16 PEs, whereas for mixed-precision GEMM, we leverage the proposed MixPE to handle INT4$\times$INT8 and INT4$\times$FP16 mpGEMM operations. 

In the process of Field Programmable Gate Array (FPGA) synthesis, various parameters can affect the circuit area and power consumption. For instance, synthesis tools can map adders and other blocks into circuits that reduce critical path delay at the cost of increased area~\cite{sklyarov2014synthesis}. To obtain the core area for a given configuration, we synthesize each configuration with easily achievable timing constraints and with only inputs and outputs being registered. This ensures that synthesis implementation selection targets the minimum area in all designs. To consider both the area and energy cost, we utilize normalized area-power product as the hardware cost. In general, to evaluate quantization accuracy, we define a statistical methodology that computes quantization SNR ratio. To measure hardware efficiency, we employ a 2D systolic array that enables the computation of synthesized power and area for different quantization configurations. Leveraging these two parameterized models, we can explore different configurations in the design space and construct a Pareto frontier that captures the optimal balance of quantization accuracy and hardware efficiency.

\section{Experiments}

\subsection{Experiment Setup}

\minisection{Accelerator Implementation}
We implement the MixPEs described in \Cref{sec:mixpe} along with traditional GEMM PEs using Verilog RTL. For synthesis, we utilized Xilinx Vivado 2023.1~\cite{Vivado} on a Xilinx Zynq UltraScale+ ZCU104 Evaluation Board, allowing us to assess resource utilization and perform static and dynamic power estimations on FPGA. For simplicity, we refer to the developed MixPE designs as ``MixPE-A8'' and ``MixPE-A16'' for W4A8 and W4A16, respectively. Additionally, we use ``INT8'' and ``FP16'' to denote the conventional INT8 and FP16 multiplication PEs. 

\minisection{Baselines}
To evaluate the end-to-end model performance of MixPEs, we compare MixPEs against four PE baselines, including conventional INT8 PE and FP16 PE, BitFusion~\cite{sharma2018bit}, and OLAccel~\cite{park2018ola}. Here we exclude ANT~\cite{guo2022ant} and Olive~\cite{guo2023olive} from comparison as both introduce additional decoders into the systolic array, incurring extra overhead. Specifically, BitFusion~\cite{sharma2018bit} supports the mixed-precision of 4-bit and 8-bit integer types. OLAccel~\cite{park2018ola} computes large values on dedicated, high-precision MAC units. We employ CACTI~\cite{muralimanohar2009cacti} to estimate the area, latency, and power consumption of memory structures. Additionally, we develop a cycle-accurate simulator inspired by DnnWeaver~\cite{sharma2016dnnweaver} and BitFusion~\cite{sharma2018bit}. We evaluate MixPE on four open-source representative LLM models, including ViT-base~\cite{alexey2020vit}, ViT-Huge~\cite{alexey2020vit}, OPT-6.7B~\cite{zhang2022opt} and LLaMA-2-13B~\cite{touvron2023llama}. These models are selected to comprehensively assess MixPE's performance across different model architectures and scales.

\minisection{Design Space Exploration}
To calculate the SNR for different quantization formats, we use the Pile dataset~\cite{gao2020pile}, sampling 128 random data points to obtain representative activation distributions. For the weight distribution, we adopt a standard normal distribution $X \sim \mathcal{N}(0, 1)$, which is commonly observed in LLM workloads~\cite{darvish2023mx}. While some layers may exhibit heavy-tailed weight distributions due to non-linear transformations (e.g., SwiGLU), addressing these variations is outside the scope of this paper. 

\subsection{Accelerator Hardware Evaluation}

As explained in \Cref{sec:mixpe}, we integrate MixPE to the systolic array-based hardware accelerator and compare its performance and energy against INT8 PE and FP16 PE adopted in commercial GPUs, such as Nvidia Tensor Core~\cite{choquette2021tensorcore} and Google TPU~\cite{jouppi2017tpu}. For a fair comparison, we set the frequency as 250MHz for all the hardware designs.
The number of columns and number of rows are both $4$, leading to a $4\times4$ systolic array structure. 

\begin{figure}
    \centering
    \includegraphics[width=0.92\linewidth]{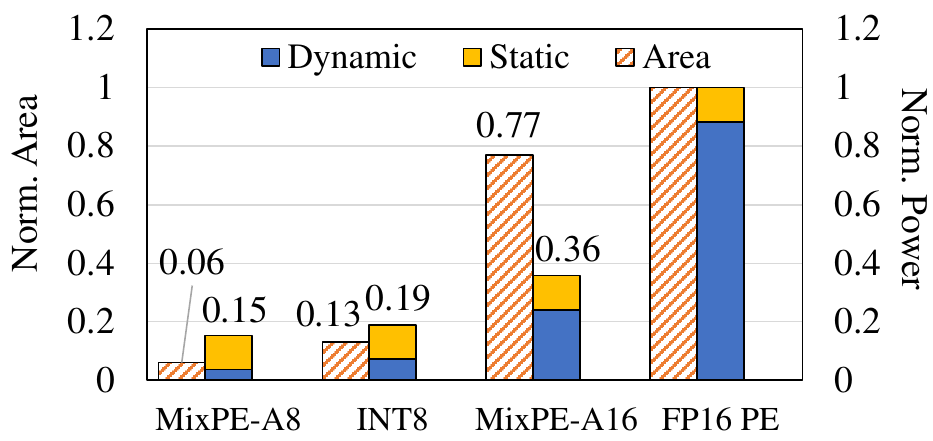}
    \caption{Comparison of normalized area and power on hardware accelerators.}
    \label{fig:area}
\end{figure}


In FPGA design, area typically refers to the amount of utilized hardware resources, including LUTs, FFs, DSPs, and IOs. The normalized area utilization across various accelerator components is illustrated in \Cref{fig:area}.
Notably, the MixPE-A8 achieves a 54\% area reduction compared to the standard INT8 PE and the MixPE-A16 reduces the area to 77\% of the FP16 counterpart, underscoring its resource efficiency. In terms of power consumption, circuit power can be divided into two primary components: dynamic and static power. The power comparison in \Cref{fig:area} reveals that the MixPE design demonstrates marked improvements in power efficiency, achieving a 64\% reduction in total power consumption compared to the FP16 PE and a 21\% reduction relative to the INT8 PE. Specifically, this power reduction mainly comes from the reduction in dynamic power, denoting that MixPE effectively minimizes switching activity by leveraging efficient shift\&add operations instead of traditional multipliers. In summary, MixPE significantly outperforms traditional multiplier-based PEs in both area and power efficiency, making it an ideal choice for mixed-precision GEMM applications on resource-constrained LLM deployments.

\subsection{Accelerator Design Space Exploration}

\begin{figure}[tb!]
    \centering
    \includegraphics[width=0.7\linewidth]{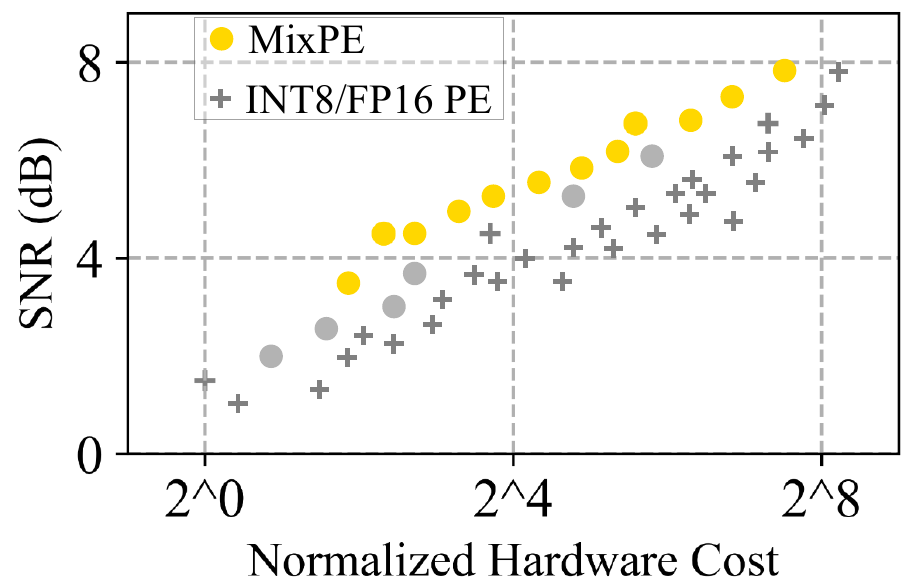}
    \caption{SNR vs.~the area-power cost for different design configurations. The gold dots denote the Pareto frontiers, achieving optimal balances between accuracy and hardware cost, while the grey dots indicate either inferior balances or low numerical fidelity.}
    \label{fig:dse}
\end{figure}

The synthesized accelerator performance, including power and area, can be affected by many aspects, such as aspect ratio, global buffer, and high-performance DSP usage. Therefore, we conducted design space exploration to gain insight in the impact of the different parameters on the performance of hardware accelerators.
\Cref{fig:dse} shows the trade-off between numerical fidelity and the hardware cost for different configurations, revealing that MixPE establishes a new Pareto frontier compared to traditional systolic array designs. In particular, for the same hardware cost, the MixPE design consistently demonstrates superior numerical fidelity, as indicated by its higher signal-to-noise ratio. This advantage stems from MixPE's highly efficient shift\&add module to minimize resource usage and energy consumption, making it a compelling choice for efficient LLM deployment in real-world scenarios.

\begin{figure*}
    \centering
    \includegraphics[width=0.99\linewidth]{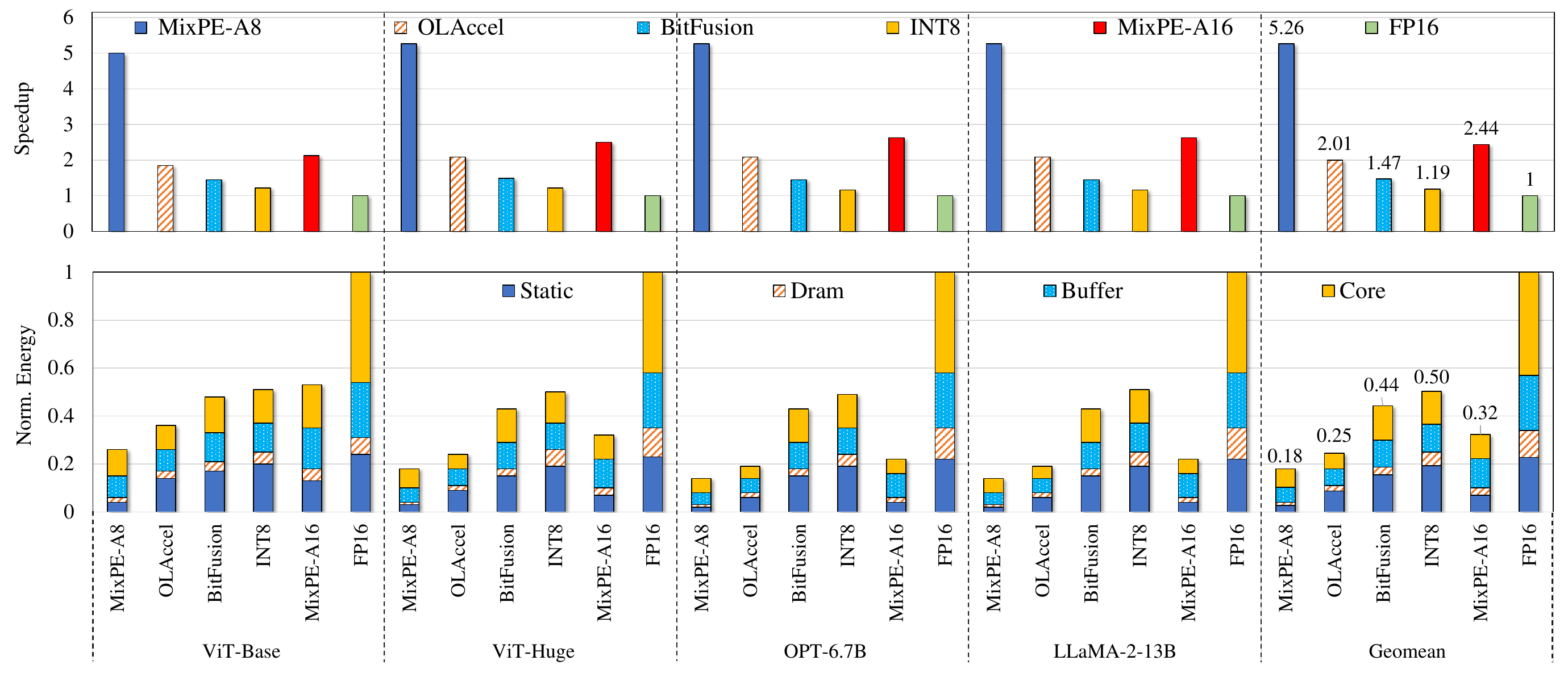}
    \caption{Comparison of speedup and normalized energy on different accelerator designs. Our MixPE design achieves 2.6$\times$ speedup and 1.39$\times$ power reduction compared to the state-of-art accelerators.}
    \label{fig:res}
\end{figure*}

\subsection{Model Performance and Energy}

We evaluate the performance of LLMs on different accelerators, with a batch size of 8. In our experiments, we compare MixPE against four baseline designs: BitFusion~\cite{sharma2018bit}, OLAccel~\cite{park2018ola}, and traditional INT8 and FP16 PE. BitFusion~\cite{sharma2018bit} utilizes a mixed-precision scheme with 4-bit and 8-bit INT types, and we extend OLAccel~\cite{park2018ola} to support Transformer-based models with 4-bit weight and 8-bit activation quantization. Note that according to the original paper, both BitFusion and OLAccel require 8-bit GEMM for certain layers. For clarity, all accelerators use the same per-group quantization scheme with a group size of 128: W4A8 for MixPE-A8, OLAccel, BitFusion, and INT8 PEs, and W4A16 for MixPE-A16 and FP16 PEs. Thus, the major difference lies in mpGEMM strategy, so the performance comparison focuses solely on accelerator performance, with metrics based on model throughput and energy efficiency. 

The normalized speedup and energy results are presented in \Cref{fig:res}. The top plot of \Cref{fig:res} demonstrates that MixPE-A8 has the most significant advantage in latency speedup. Specifically, MixPE-A8 delivers speedups of 2.62$\times$, 3.58$\times$, and 4.42$\times$ over OLAccel, BitFusion and INT8, respectively. MixPE-A16, on the other hand, achieves a 2.44$\times$ speedup over its FP16 counterpart and even outperforms the other W4A8 baselines, i.e., BitFusion, OLAccel, and INT8, due to the substantial cycle reduction enabled by the dequantization-after-GEMM strategy. Meanwhile, the speedup values for MixPE are consistent across different models, owing to the similarity in the decoder-only architectures of the language models, with differences mainly in the number of layers and hidden state parameters.

The bottom plot of \Cref{fig:res} compares the normalized energy consumption of different designs, which includes the static energy and dynamic power (DRAM, on-chip buffer, and core). MixPE-A8 exhibits the lowest energy consumption, achieving an average reduction of 2.21$\times$ compared to the other designs. Specifically, MixPE-A8 achieves a power reduction of 1.39$\times$, 2.44$\times$, and 2.78$\times$ over OLAccel, BitFusion, and INT8, respectively. Additionally, MixPE-A16 decreases the energy consumption of FP16 by 68\% on average. Notably, the energy reduction becomes more pronounced as the model size increases. For instance, MixPE-A16’s energy consumption is even lower than that of lower-bit BitFusion and INT8 on the three larger language models, highlighting its scalability. This enhanced energy efficiency is attributed to the optimized shift\&add modules in MixPE-A16, which leverage low-bit precision data to maximize computational efficiency. Overall, by co-optimizing both the quantization algorithm—specifically the dequantization-after-GEMM strategy—and the hardware accelerator, i.e., the shift\&add PE design, MixPE fully exploits the benefits of low-bit precision quantization, leading to substantial reductions in both energy and computational overhead.

\subsection{Ablation Study of Dequantization Overhead}

\begin{figure}
    \centering
    \includegraphics[width=0.93\linewidth]{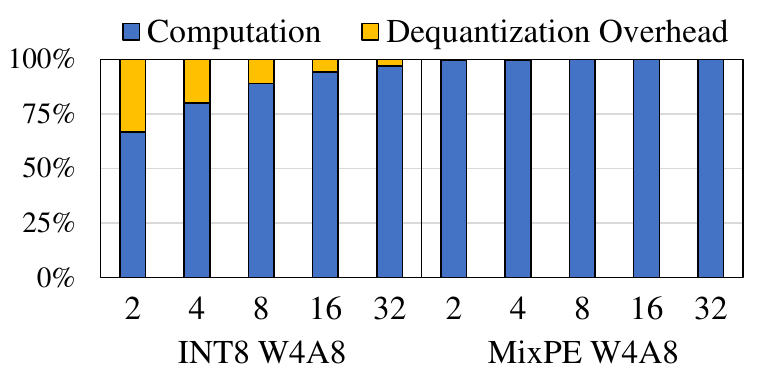}
    \caption{The W4A8 dequantization overhead on MixPE and INT8 PE. The dequantization overhead on MixPE is much smaller than that in INT8 PE.}
    \label{fig:dequant}
\end{figure}

We measure the dequantization overhead of the mixed-precision quantization algorithm on both MixPE and INT8 PE. Our analysis focuses on the W4A8 quantization scheme applied to OPT-6.7B, with batch sizes ranging from 2 to 32. The dequantization overhead for both MixPE and INT8 PE is presented in \Cref{fig:dequant}. The results indicate that MixPE exhibits significantly lower dequantization overhead compared to the INT8 PE. This reduction becomes more pronounced at lower batch sizes due to the relatively smaller computational load. In practical LLM deployment settings, where batch sizes are often small but the sequence length is long, MixPE emerges as the more efficient choice, offering substantial improvements in both energy and computational efficiency during real-world inference tasks.

\section{Conclusion}

In this work, we present MixPE, a novel hardware accelerator designed to efficiently handle mixed-precision GEMM with minimal dequantization overhead. The key insight lies in exploiting per-group parameter sharing to enable dequantization after mpGEMM, thereby reducing overhead in the main loop. To fully capitalize on the benefits of low-precision weights, we propose a dedicated shift\&add-based processing element to achieve mpGEMM without energy-intensive multipliers. MixPE pushes the efficiency of LLM quantization to a new level by co-designing the quantization scheme and hardware accelerator. Moreover, we conduct an exhaustive exploration of design variables, demonstrating that MixPE establishes a Pareto frontier that optimally balances numerical fidelity and hardware efficiency. Finally, MixPE outperforms existing accelerators, achieving a $2.6\times$ speedup and $1.4\times$ reduction in energy consumption.

{
\bibliographystyle{IEEEtran-sim}
\bibliography{reference}
}

\end{document}